\documentclass[10pt,twocolumn,letterpaper]{article}

\usepackage{wacv}
\usepackage{times}
\usepackage{epsfig}
\usepackage{graphicx}
\usepackage{amsmath}
\usepackage{amssymb}

\wacvfinalcopy 

\def\wacvPaperID{****} 

\ifwacvfinal\fi
\begin{document}

\title{Super-Resolution for Overhead Imagery Using DenseNets and Adversarial Learning}

\author{Marc Bosch \\
The Johns Hopkins University \\
Applied Physics Laboratory\\
{\tt\small marc.bosch.ruiz@jhuapl.edu}
\and
Christopher M. Gifford \\
The Johns Hopkins University\\
Applied Physics Laboratory \\
{\tt\small Christopher.Gifford@jhuapl.edu}
\and
Pedro A. Rodriguez \\
The Johns Hopkins University\\
Applied Physics Laboratory\\
{\tt\small Pedro.Rodriguez@jhuapl.edu}
}

\maketitle
\ifwacvfinal\thispagestyle{empty}\fi

\begin{abstract}
 Recent advances in Generative Adversarial Learning allow for new modalities of image super-resolution by learning low to high resolution mappings. In this paper we present our work using Generative Adversarial Networks (GANs) with applications to overhead and satellite imagery. We have experimented with several state-of-the-art architectures. We propose a GAN-based architecture using densely connected convolutional neural networks (DenseNets) to be able to super-resolve overhead imagery with a factor of up to $8\times$. We have also investigated resolution limits of these networks. We report results on several publicly available datasets, including \textit{SpaceNet} data and \textit{IARPA Multi-View Stereo Challenge}, and compare performance with other state-of-the-art architectures. 
\end{abstract}

\section{Introduction}
Super-resolution is the task of estimating plausible pixel information given an image and creating a corresponding higher resolution version. Typically, super-resolution methods aim to recover high frequency components of the scene lost in the image acquisition process for an improved perceived quality. The majority of approaches attempt to either produce new pixel values by estimating them from a support region (neighborhood) or through a learned model given many examples of low resolution to high resolution mappings. The latter has been an active area of research for many years within the computer vision and image processing communities~\cite{D-bib:freeman2, keys, yang, schutler, srcnn}. Some of these works have proven successful for recovering detail from low resolution images typically acquired with consumer electronic cameras. However, the low resolution input of many available solutions still offer enough detail to infer most of the semantics of the scene.
\begin{figure}[ht]
   \includegraphics[scale=0.5]{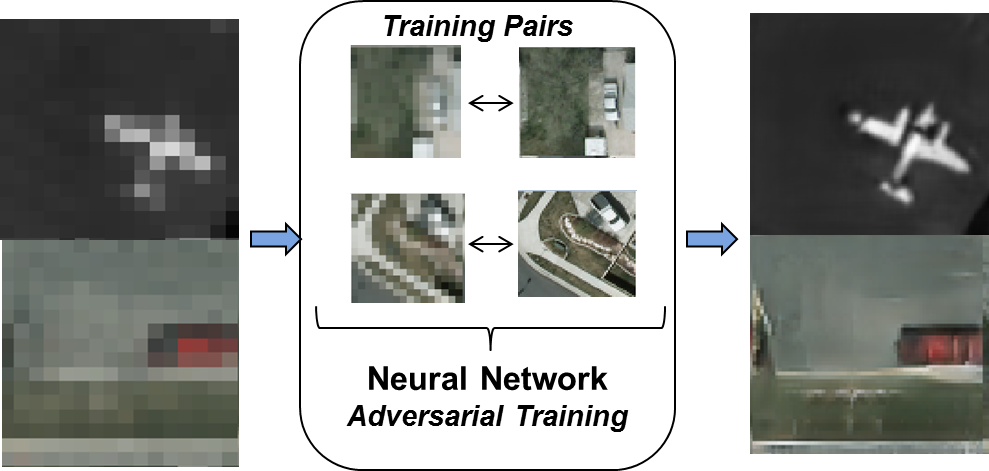}
   \caption{Super-resolution applied to overhead imagery using our system.}
\label{fig:example_deepres}
\end{figure}

In overhead imaging (\textit{i.e.} airborne and satellite), super-resolution has the potential to offer advanced automatic target recognition (ATR) capabilities and improved human exploitation value. Due to the distance from source to target, even a small resolution gain can dramatically improve the end use. Super-resolving an image by an upscale factor of 4 means that from one pixel we need to estimate fifteen new pixels in a $4 \times 4$ neighborhood. In other words, if the original image resolution represents a ground sampling distance (GSD) of four meters per pixel, the new super-resolved image would resolved up to one meter per pixel. In this context, many common objects would not have enough detail for analysts to ``understand'' the scene. Hence, it becomes more a problem of enhancing semantics, objects, and fine-grained features rather than improve image quality.

Super-resolution of overhead imagery can have a significant impact for the space industry. There are many potential remote sensing applications that would directly benefit from this technology, such as crops and deforestation monitoring, economic activity tracking, space imaging, and various reconnaissance activities. Also, different imagery modalities such as panchromatic electro-optical (EO), hyperspectral (HSI), infrared (IR), and synthetic aperture radar (SAR) can benefit from these advances.

With the increasing attention being given to neural network models, recent works have proposed the use of several network architectures to tackle the super-resolution problem, including~\cite{srcnn, srgan, pixelsrnn}. Generative models have been proposed to infer and recover plausible details from low resolution images. Generative Adversarial Networks (GANs) are one of the most popular generative Deep Learning framework for super-resolution. GANs are trained through adversarial training where two engines, namely \textit{generator} and \textit{discriminator}, participate in a two-player game with the goal of improving as the opponent improves. In ~\cite{srgan}, authors proposed a GAN-based algorithm using neural networks with deep convolution layers to model the low-to-high resolution mappings.

In this work, we have investigated multiple convolutional neural network (CNN) configurations. Further, we propose the integration of a densely connected convolutional network architecture (DenseNet) within the \textit{generator} of a GAN. DenseNets~\cite{densenet} have the particularity of connecting layers in a dense manner to other layers for better feature representation and computational efficiency.  Figure~\ref{fig:example_deepres} shows an example of inputs/outputs of our framework applied to satellite images.

\subsection{Background}
Work in super-resolution has been ongoing for decades now. There have been two main approaches: multi-frame and single frame super-resolution. Among single frame/image super-resolution, we can find baseline algorithms like bicubic interpolation, cubic splines~\cite{keys}, and other local-based approaches that consider local regions of support to estimate new pixels. Some techniques rely on statistical priors from the image~\cite{D-bib:freeman2}. More recently, learning-based approaches have successfully been used to produce improved results. These techniques are based on the self-similarity principle, where it is assumed that many images share similar visual properties at scale. These approaches back-project high-frequency components lost in the low resolution image from similar patches found in other images given a large collection of reference images~\cite{schutler, D-bib:freeman3, yang, Zeyde}.

As in many other applications in computer vision, Deep Learning techniques have prevailed in the last few years. There have been several proposed systems where the non-linear mapping between low and high resolution images is learned end-to-end using neural networks~\cite{srcnn, srgan, pixelsrnn, laplacian, tai}.

Dong et al. proposed the use of Deep CNNs (SRCNN) to learn the high-frequency representation given low resolutions similar to sparse-coding-based techniques with the added advantage of the joint optimization that occurs in a Deep Learning system~\cite{srcnn}.

In~\cite{drcn}, authors proposed a very deep network that exploits the advantages of a recursive architecture like adding convolutions without adding new parameters for improved performance, and also adding skip connections to reduce the effect of the vanishing gradient found in the recursive schemes.

Dahl and colleagues proposed a method that produced very impressive results for the task of super-resolving human faces~\cite{pixelsrnn}. Their method is based on a network that operates in a recursive fashion. Each pixel is super-resolved using  previously visited pixels. They combine two CNNs, one based on PixelCNN that makes predictions using previous stochastic estimates, and another that behaves as the conditioning network by receiving the low resolution image and generating logits that encode the log-probability of each pixel in the high resolution image.

In~\cite{bruna}, authors proposed a method that uses CNNs to characterize high-frequency content of images not present in the source image. They model the conditional mapping of a high-resolution image given its low-resolution version as a Gibbs distribution. Following this work, Ledig et al., in~\cite{srgan}, proposed the use of adversarial training and GANs to tackle the recovery of high-level details of an image. They designed a framework that uses residual blocks~\cite{resnet} to generate features that encode plausible missing information. The weights of the network are updated so that the adversarial loss combined with feature matching loss is minimized. They add a feature matching loss to quantify the fidelity of the reconstructed image in terms of perceptual loss by a VGG-19 universal feature extractor.

\subsection{Contributions}
In this paper, we describe the integration of densely connected convolutional networks to a GAN framework for the task of super-resolution for overhead imagery. Our main contributions can be summarized as follows:
\begin{itemize}
\item Investigated application of existing state-of-the-art super-resolution models to overhead imagery to determine what is possible with today's models for both panchromatic electro-optical (EO) and multi-band images.
\item Proposed a network architecture based on dense blocks repurposed for super-resolution applications under an adversarial training framework. 
\item Evaluated several loss functions, including feature matching criteria, with the purpose of understanding how transferable pre-trained models are on non-overhead natural images with much different scale and geometry viewpoints with respect to satellite imagery.
\item Evaluated several super-resolution gain factors to understand the limits of these techniques, as well as constrain the scope of the problem by conducting experiments on specific semantic categories. We have used imagery with a large diversity of geometric and semantic features to gain an understanding on the limits and capabilities of this technology.
\end{itemize}

The remainder of the paper is organized as follows: In Section 2 we review the GAN model, in Section 3 we present our proposed method using a DenseNet, in Section 4 we show our experimental results on overhead imagery, and finally, we conclude the paper by offering some remarks from our experimental observations.
\section{Generative Adversarial Network (GAN) Models}
Generative Adversarial Networks (GANs) are a particular case of generative models. Their goal is to learn the probability distribution of the source data using adversarial training. This means applying alternating stochastic gradient updates in a two-player, zero-loss game. In the task of super-resolution, a generative model is required to input new details into the low resolution input. Two engines, \textit{generator} and \textit{discriminator}, are established in a min-max optimization framework. The \textit{generator} aims at generating new samples, \textit{fake data}, from the learned data distribution that look as real as possible. The \textit{discriminator}'s goal is to detect such \textit{fake data} among a collection of real training data. At training time, both \textit{generator} and \textit{discriminator} take turns to fool the opponent and to distinguish the fake from real data respectively. As a result, training is successful if the \textit{generator} becomes increasingly better at creating realistic data, and the \textit{discriminator} improves spoofing detection capabilities, which forces the \textit{generator} to become even better in its attempt to blur the line between \textit{fake} and \textit{real} data.

More formally, the adversarial training process in a GAN is described by the following adversarial cost/loss function:
\begin{equation}
\label{eq:lossadv}
\begin{split}
Loss_{G,D}=amin_{\psi_G}max_{\psi_D}(E_{x\sim p_{target}(x)}(logD_{\psi_D}(x))+\\
E_{z\sim p_{model}(z)}(log(1-D_{\psi_D}(G_{\psi_G}(z))))
\end{split}
\end{equation}
with $x$ being samples from the target distribution (i.e., high-resolution training images), and $z$ representing the input variables needed in the model estimate (i.e., low-resolution training images) to generate new high-resolution images that approximate to the target high res images $x$. During training, the \textit{discriminator} maximizes the following expression given a batch of generated data from the \textit{generator}, $G_0(z)$:
\begin{equation}
\begin{split}
\psi_D^{(*)}=max_{\psi_D}(E_{x\sim p_{target}(x)}(logD_{\psi_D}(x))+ \\
E_{z\sim p_{model}(z)}(log(1-D_{\psi_D}(G_{\psi_{G_0}}(z)))))
\end{split}
\end{equation}
Next, the \textit{generator} objective is to minimize the following:
\begin{equation}
\psi_G^{(*)}=argmin_{\psi_G} E_{z\sim p_{model}(z)}(log(1-D^{(*)}_{\psi_D^{(*)}}(G_{\psi_G}(z))))
\end{equation}

\paragraph{Deep Convolutional GAN}
Deep Convolutional Generative Adversarial Networks (DCGANs) are a class of GANs implemented, in general, with several convolutional layers. Both the generator and discriminator can be implemented using CNNs and applied to several tasks, such as image generation (generator) or image classification (discriminator)~\cite{dcgan, lapgan-denton}. Two CNN architecture types are found in most of the work using DC-GANs, these are encoder-decoder type architecture and ResNet~\cite{resnet}. Note that both do not necessarily need to be complementary.

In the encoder-decoder approach, the input is passed through a series of layers that downsample the feature maps to be able to represent larger receptive fields. A decoder brings back the information to the full input resolution.

ResNet is made of residual blocks that, at each layer, learn a ``residual'' mapping, $R(x)$, obtained from subtracting the underlying input ($x$) to output ($y$) mapping, $H(x)$, from the layer input, $y=H(x)=R(x)+x$. These residual mappings are learned and added to the identity mapping (shortcut). ResNet has been proven to outperform other state-of-the-art approaches in many tasks, including object recognition. Shortcut connections allow the network to skip certain layers to avoid vanishing gradients and curse of dimensionality problems.
\section{Dense Network GANs for Super-Resolution}
In~\cite{densenet}, the authors introduced a new type of CNN architecture, DenseNet, that allows for a richer description of the visual elements in the scene compared to state-of-the-art networks like ResNet. In short, a dense network is a collection of blocks where each of the block's layers is connected to one another; hence, they are referred to as \textit{dense} blocks~\cite{densenet}.

Dense blocks have the advantage over residual blocks in that there is a stronger gradient flow due to direct connection at any layer from all other layers in the network. It also maintains and exposes low complexity features at deeper layers in the network better than ResNet does. These much richer representations of visual attributes enable better parameter and computational efficiency, which better positions DenseNet for onboard processing in airborne and spaceborne platforms.
\subsection{DenseNet Generator}
\label{ssec:generator}
Our implementation of a DenseNet is limited to the generator of a GAN. It consists of a dense block with BN-ReLU-Conv ($3\times 3$) basic unit sequence, preceded by a bottleneck block to reduce the number of input feature maps composed by the BN-ReLU-Conv ($1\times 1$) layer. In addition, each dense block is separated from its neighboring (dense) block by a transition layer. In the original dense network, the transition layer is made of a convolution layer and a pooling layer. In our super-resolution framework, we modify this layer to account for the resolution increase goal. Rather than performing all the convolutions in the low resolution domain as other approaches do, the high efficiency dense blocks allow us to learn the mappings at several scales. Our transition layer doubles the resolution of the feature maps each time. This in turn introduces smaller receptive fields that learn local information, and, thus, recovers fine-grain details. One obvious benefit of this cascaded process is that the network is trained at several resolution gains, making it a more versatile design.

As shown in~\cite{fcn}, a way to connect coarse outputs to dense pixels is using backward convolutions, also known as deconvolutions or transpose convolutions. We use deconvolution with a $stride=2$ to generate an upscale version ($2\times$) of the input feature map at each transition layer.

\begin{figure}[ht]
   \includegraphics[scale=0.45]{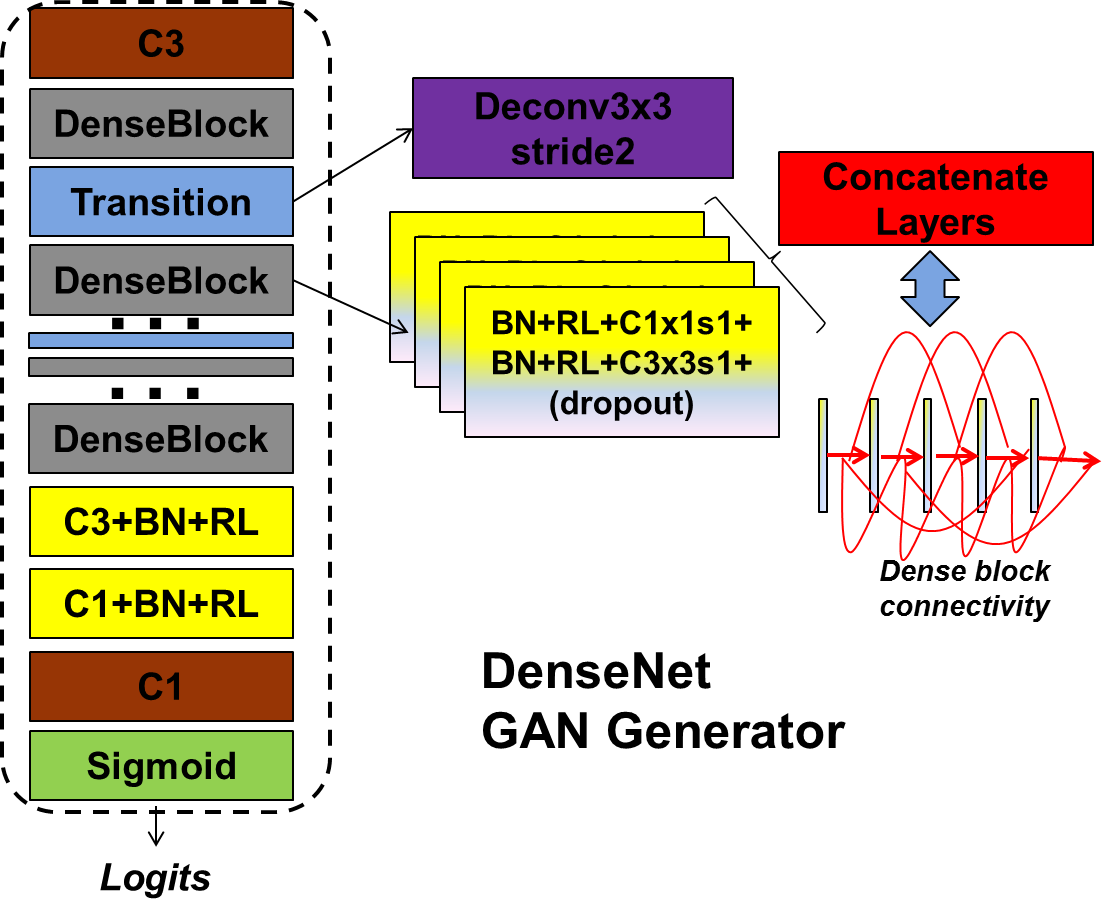}
   \caption{Generator architecture for overhead imagery.}
\label{fig:denseSRGAN}
\end{figure}

Following~\cite{tobias, david-gpusrez} implementations, the last stage of the generator is a fully-convolutional network.
Figure~\ref{fig:denseSRGAN} shows our proposed generator architecture for super-resolution.

\subsection{Discriminator}
\begin{figure}[ht]
\begin{center}
   \includegraphics[scale=0.55]{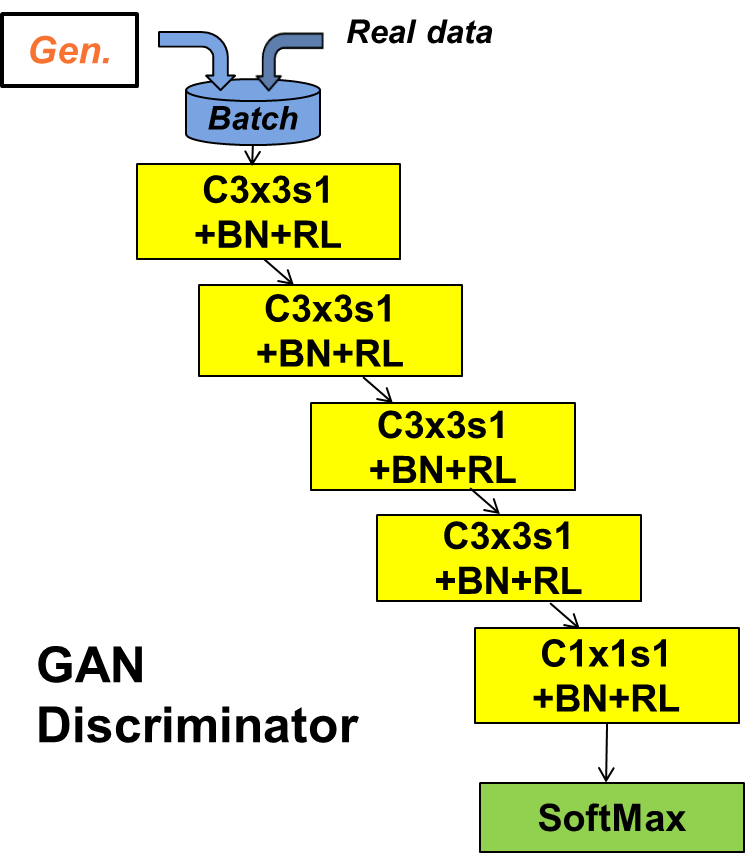}
\end{center}
   \caption{Discriminator architecture for training the generator.}
\label{fig:discr}
\end{figure}

Following other discriminatory models, our discriminator has a relatively shallow configuration with a series of convolutional layers followed by ReLUs and batch normalization blocks. At each layer, the convolutional layer is doubled in a similar fashion as the VGG network~\cite{srgan}. Again, a fully-convolutional subnet is placed in the final layers~\cite{tobias, david-gpusrez}, followed by sigmoid functions that output the decision of real vs. fake input data. Figure~\ref{fig:discr} shows the architecture of the discriminator used to train the generator.
\subsection{Loss Function}
In this work, we have investigated several generator loss types to guide the network to convergence. In general form, the loss can be expressed as a function of the adversarial loss, L1-norm image content loss, and feature matching loss.
\begin{equation}
\begin{split}
Loss_{Gen}=\alpha(n) Loss_{adv}+\\
(1-\alpha(n))((1-\beta_1) \cdot Loss_{content}+\\
\beta_1 \cdot Loss_{fm})
\end{split}
\end{equation}
The $Loss_{adv}$ is the ``vanilla'' GAN adversarial loss:
\begin{equation}
Loss_{adv}=E_{z\sim p_{model}(z)}(log(1-D_{\psi_D}(G_{\psi_G}(z))))
\end{equation}
$Loss_{content}$ represents the L1-norm between the target high-resolution and the generated high-resolution images:
\begin{equation}
Loss_{content}=\| targetHR-gen.HR \|_1
\end{equation}
Similarly, the feature matching loss ($Loss_{fm}$) has the goal of describing the visual attribute loss. We use a version of the popular VGG-16 using pre-trained weights on ImageNet as a universal feature extractor. It is computed as:
\begin{equation}
Loss_{fm}=\| T_{\theta_{vgg16}}(targetHR)-T_{\theta_{vgg16}}(gen.HR) \|_1
\end{equation}
The parameter $\alpha$ dynamically modulates the influence of the adversarial loss into the overall loss. Similarly, $\beta_1$ controls the importance of the feature matching loss.  
The discriminator only uses the adversarial loss, eq.~\ref{eq:lossadv}. It is simply a binary cross entropy loss.
\subsection{Implementation and Training Details}
The generator of the proposed GAN using dense blocks has 5 dense blocks per layer. Each block consists of a bottleneck with a batch normalization block, a ReLU and a convolutional layer with $1\times 1$ map size (filter spatial support) and $stride=1$, followed by a combination of batch normalization, ReLU and convolutional layer, with a $3\times 3$ map size and $stride=1$. At each of these units, we add 16 feature maps. This is also referred to as the growth rate of the network. The growth rate is kept small to avoid the network from growing too wide and to improve parameter efficiency. The transition layer consists of a deconvolution unit with a $3\times 3$ and $stride=2$ to account for the upscale.

As mentioned earlier in Section~\ref{ssec:generator}, the final layers of the generator are an implementation of the fully-convolutional network~\cite{tobias}, in particular a layer with convolutional filters with a mapsize of $3\times 3$ with $stride=1$ followed by batch normalization, and ReLUs. The block is repeated once more but with a bottleneck convolution stage of $1\times 1$ size. In this final stage, the size of the tensor is not changed as no new feature maps are added.

The discriminator is a four layer network with [64, 128, 256, 512] feature maps for the respective layers. Each filter in each of the layers has $3\times 3$ filter size and $stride=2$. The fully-convolutional network from the generator is mimicked after the four layer stages.

In our reported results, our settings are as follows: Batch size is 16. We initialize the parameter controlling the contribution of each loss type, $alpha$, to $0.95$ to guide the network towards perceptual convergence at the initial stages, and decrease at each epoch by a factor of $1.05$. 
Finally, we have evaluated the network for several settings of $beta$ to investigate the contribution of each term of the perceptual loss, namely feature matching loss and the pixel content loss. Our goal was to evaluate the correlation of networks trained to extract features trained on non-overhead non-synthetic images like VGG-16 with overhead imagery with such a different viewing geometry.

\section{Evaluation}
In this section, we describe our experiments and present results for image super-resolution for overhead imagery. We compare several methods using an objective quality metric, PSNR, as well as show visual results of the proposed method. For all intents and purposes, we have simulated sensor resolution limitations from overhead imagery as a nearest neighbor down-sampling model.
\subsection{Satellite Imagery Datasets}
\begin{figure}[ht]
	\includegraphics[scale=0.65]{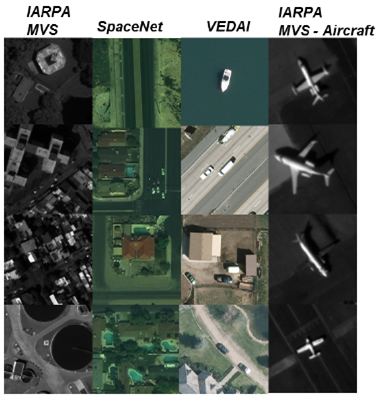}
   \caption{Example samples used in our overhead imagery experiments.}
\label{fig:datasets}
\end{figure}

We have conducted experiments on several public overhead imagery datasets. These include \textit{SpaceNet Challenge}~\cite{spacenet}, the \textit{IARPA Multi-View Stereo Satellite Challenge}~\cite{iarpamvs}, and the Vehicle Detection in Aerial Imagery dataset (\textit{VEDAI})~\cite{vedai}. See figure~\ref{fig:datasets} for examples of each.

The SpaceNet dataset was released as part of the \textit{SpaceNet} Challenge. In our experiments, we have used the multi-band images corresponding to the AOI-2 site, which captures several views of the city of Las Vegas. These images have 30cm GSD and were collected with the WorldView-3 sensor from Digital Globe~\cite{spacenet}. 
Another dataset used was gathered from the recent \textit{IARPA Multi-View Stereo (MVS) Challenge}. This dataset consists of 50 WorldView-3 panchromatic images with 30cm GSD over a 50 sqkm area near Buenos Aires in Argentina ~\cite{iarpamvs}. We have also extracted chips with airplanes from this imagery to evaluate the algorithms with targeted features. 
The \textit{VEDAI} dataset was released in 2015 as a benchmark for vehicle detection tasks in aerial imagery. It has more than a thousand images with various objects of interest, including vehicles, boats, tractors, and aircraft. 
Table~\ref{tab:datasets} summarizes the data used in our experiments.
\begin{table}[ht]
\caption{Summary of Datasets.}
\centering
\begin{tabular}{c c c}
\hline\hline
 Dataset & Chip Size & Number of Image Chips \\
 \hline
 SpaceNet - & & \\
 Las Vegas  & $256\times 256$  & 45266 \\
 \hline
 IARPA MVS - & & \\
 All  & $256\times 256$  & 382795 \\
 \hline
 IARPA MVS - & & \\
 Aircraft  & $344\times 344$  & 1056 \\
 \hline
 VEDAI  & $256\times 256$  & 3734 \\
 \hline
\hline
\end{tabular}
\label{tab:datasets}
\end{table}
\subsection{Super-Resolution Results}
\begin{figure*}[h]
	 \includegraphics[scale=0.45]{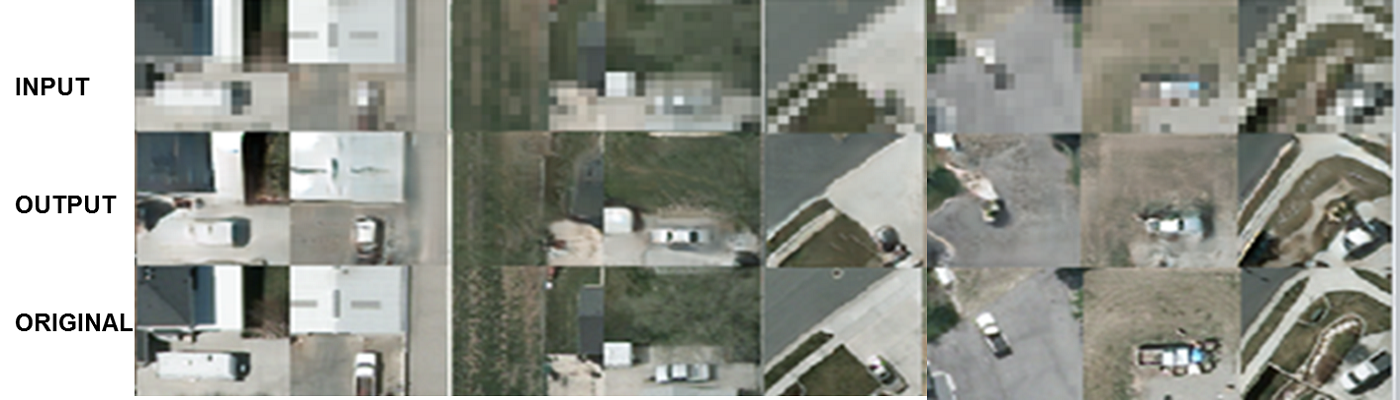}
   \caption{Results on the VEDAI dataset. Super-resolution factor of 4.}
\label{fig:vedai}
\end{figure*}

\begin{figure*}[h]
	 \includegraphics[scale=0.45]{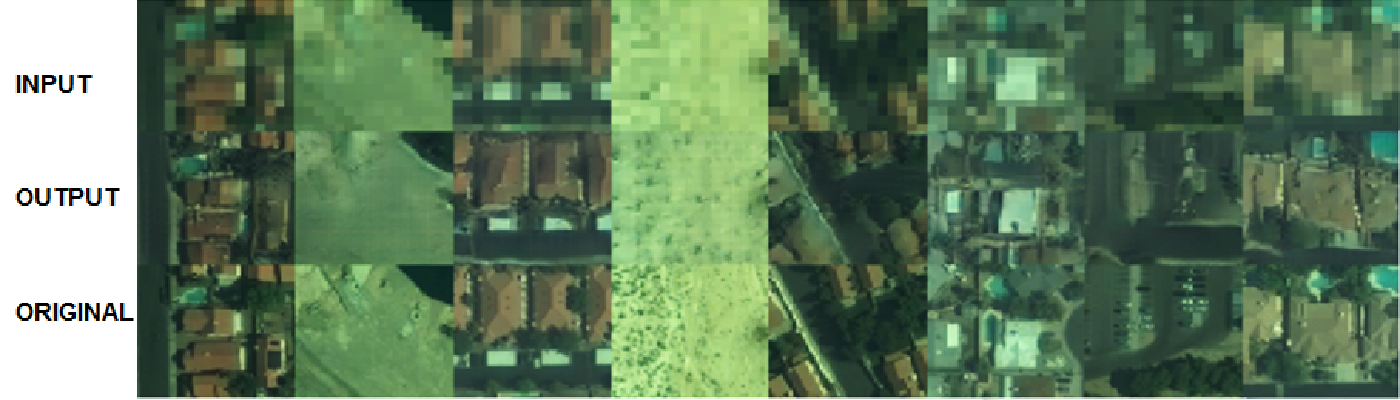}
   \caption{Results on the SpaceNet Las Vegas dataset. Super-resolution factor of 4.}
\label{fig:vegas}
\end{figure*}

\begin{figure*}[ht]
	 \includegraphics[scale=0.45]{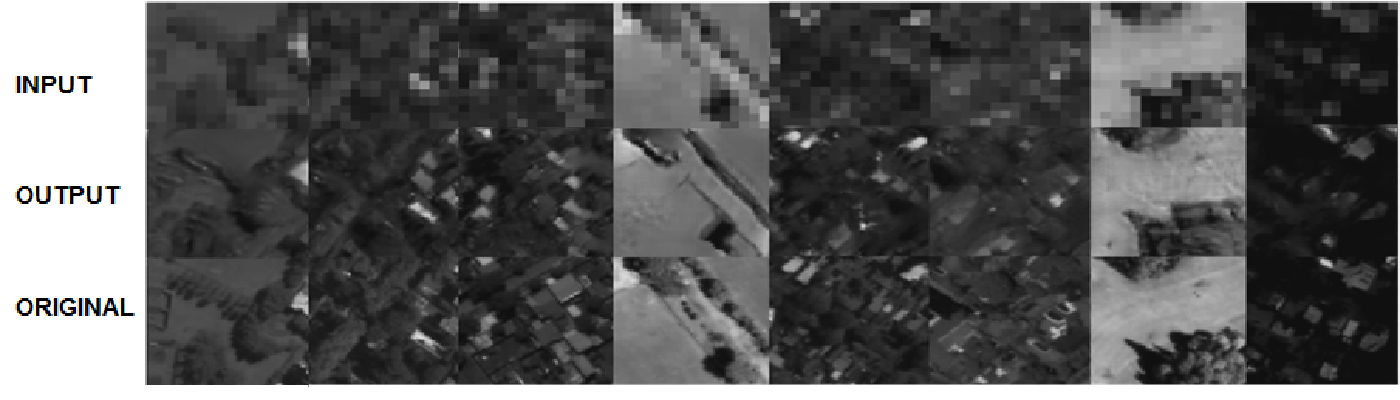}
   \caption{Results on the IARPA MVS dataset. Super-resolution factor of 4.}
\label{fig:iarpaarg}
\end{figure*}

We have conducted our experiments on several overhead images, as described in the previous section. Results are reported using PSNR on a validation subset of images for each dataset. Table~\ref{tab:results} summarizes a comparison of our approach with several state-of-the-art algorithms based on other CNN architectures. We have focused our analysis and comparisons on two GAN-based models. First, a system proposed by Ledig et al., in~\cite{srgan}, one of the pioneering works on using GANs for super-resolution. We refer to this work as \textit{SR-DCGAN}. Second, an approach introduced in~\cite{pix2pix} to tackle the image translation problem has been repurposed for super-resolution tasks in our work. Note that this work was not explicitly defined for super-resolution, but rather for a more generic set of image domain mappings. Finally, the last algorithm we have used for comparisons was presented in~\cite{pixelsrnn}. It is a non-GAN scheme based on PixelCNN that uses pixel recurrency to predict current samples.
\begin{table}[ht]
\small
\caption{Comparison between state-of-the-art network architectures for super-resolution tasks for three overhead imagery datasets. (L1: L1-loss, FM: Feature matching loss, A: Adversarial loss)}
\centering
\begin{tabular}{c c c c c}
\hline\hline
 Algorithm & VEDAI & SpaceNet & IARPA & Average\\
  & & Vegas & MVS & \\
           & Quality & Quality & Quality & Quality \\
					& dBs & dBs & dBs & dBs \\
 \hline
 SR-DCGAN & 29.4 & 29.6 & 28.3 & 29.1\\
~\cite{srgan} & & & & \\ 
 \hline
 PixelCNN  & 27.7 & 29.1 & 28.3 & 28.4\\
~\cite{pixelsrnn} & & & & \\
 \hline
 pix2pix & 29.0 & 30.8 & \textbf{29.9} & 29.9\\
~\cite{pix2pix} & & & & \\
\hline
  DenseNet & & & & \\
  GAN & & & & \\
  loss: L1, & 29.1 & 30.7 & 27.5 & 29.1\\
	FM, A (Ours) & & & & \\
 \hline
  DenseNet & & & & \\
	GAN & & & &\\
  loss: L1, A & \textbf{29.9} & \textbf{31.3} & 29.6 & \textbf{30.3}\\
	(Ours) & & & & \\ 
 \hline \hline
\end{tabular}
\label{tab:results}
\end{table}

In figures~\ref{fig:vedai},~\ref{fig:vegas}, and~\ref{fig:iarpaarg} we show results of the proposed GAN scheme using dense blocks applied to \textit{VEDAI}, \textit{SpaceNet - Las Vegas}, and \textit{IARPA MVS} datasets for a factor of $4\times$ super-resolution. These results present a more tangible description of the true performance of our system for overhead imagery.
\begin{figure}[ht]
\begin{center}
   \includegraphics[scale=0.55]{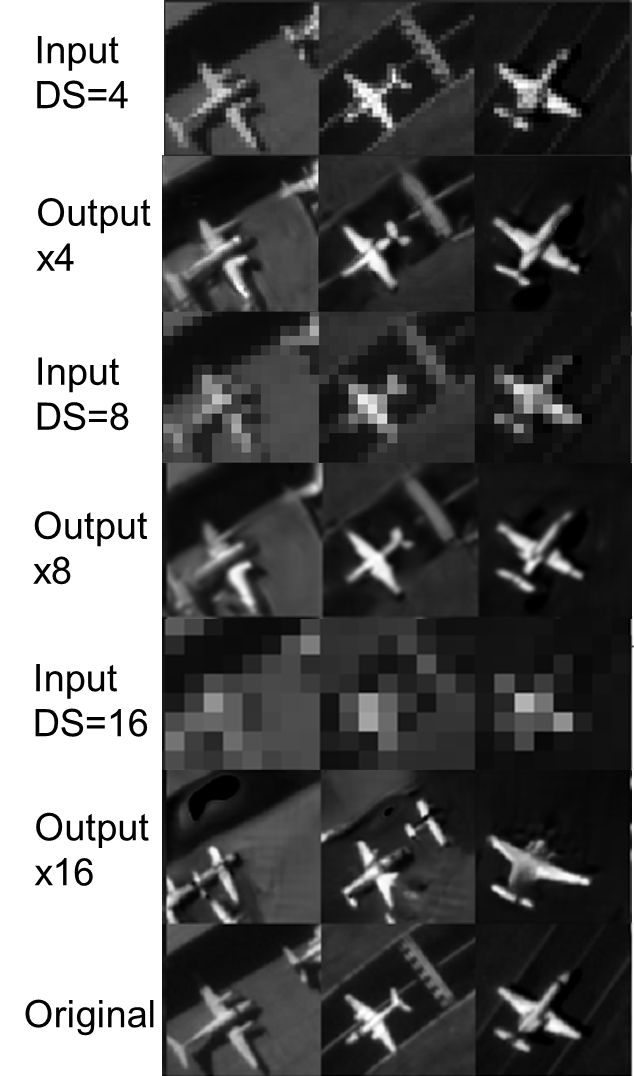}
   \caption{Results of the proposed system as the resolution factor is doubled.}
\label{fig:planesscale}
\end{center}
\end{figure}
\paragraph{Limits of Extreme Super-Resolution}
We also looked into super-resolution transformations at larger scale factors, \textit{e.g.,} $>4\times$, as well as identifying limits beyond which these systems become unsuited for processing. Figure~\ref{fig:planesscale} shows the degradation as we double the resolution factor, or in other words as we downsample the input while aiming for same output resolution.

\begin{figure}[ht]
\begin{center}
   \includegraphics[scale=0.5]{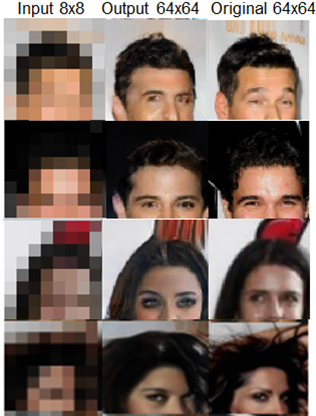}
   \caption{Results of the proposed system on human faces.}
\label{fig:faces}
\end{center}
\end{figure}

Finally, for reference we provide results on non-overhead imagery. We used the popular \textit{CelebA} dataset containing many examples of human faces. Figure~\ref{fig:faces} shows some results of our proposed architecture using DenseNet in the generator. Images have been super-resolved by a factor of 8, \textit{i.e.,} from $8\times 8$ inputs to $64\times 64$ outputs.

\section{Results Discussion}
Mapping from lower resolution images to higher resolutions is a specific case of the broader image translation problem. Fine-grained scene details can be lost during the image acquisition process. In overhead imagery, and long-range imaging applications, this translates into information, thus intelligence, loss. Our goal is to produce new plausible information to limit the impact of the imaging system resolution loss for lower cost imagers and long-range imaging applications. We have designed a system that is shown many examples of low-resolution and high-resolution image pairs and asked to learn the non-linear mapping that occurs.

From the results we have shown, we can see that overhead imaging has its own set of challenges (\textit{e.g.,} there is a large variation of features present in the scene). It is difficult to obtain a well-balanced training set that allows to fully model the different viewing geometries of the sensor and variety of objects and geospatially diverse backgrounds present in various scenes. Also, having a nadir view constrains the problem somewhat, as we can see by comparing the results obtained for the \textit{SpaceNet} dataset, figure~\ref{fig:vegas}, in contrast to results for the \textit{IARPA MVS} collection with much more angular diversity (figure~\ref{fig:iarpaarg}). The proposed super-resolution algorithm can recover information better in the former case than the latter case.

Comparing the results of the proposed method with the other algorithms (see table~\ref{tab:results}), we can see that adding layers of DenseNet blocks into the GAN generator improves the performance of other configurations for this task. These results are aligned with findings and claims found in~\cite{densenet}, where dense blocks are superior schemes compared to more traditional convolutional and residual layers due to being able to expose lower-level features at deeper layers of the network. This is certainly an advantage for low-level tasks such as super-resolution. We did not observe any improvement by adding more dense blocks or increasing the growth rate. We tried to keep the complexity of the network as low as possible with no noticeable loss in performance.

As perhaps expected, the network performed worse when using VGG-16 pre-trained on Imagenet images as a universal feature extractor to compute the feature matching loss component of the overall loss function, as shown in table~\ref{tab:results} (last two rows). There is little to no correlation between features found in a natural image to the features found in an overhead image. A standard universal feature extractor does not efficiently capture visual attributes at such different geometric viewpoints.

Figure~\ref{fig:planesscale} shows the performance of the network when resolution of the input image is reduced further. At $8\times$ factors (fourth row in the figure) it is still capable to recover and properly estimate the content in the image. However, at lower input resolutions (second from the bottom row) the network is completely ``hallucinating'' the wrong content. These types of exercises are useful, though, to show limitations, feasibility, and generalization of the solution.

We have also observed that, when training data is constrained to particular semantic categories (\textit{e.g.,} aicraft dataset extracted from the \textit{IARPA MVS} images), the capability of the network to train the probability distribution is quite good even with a small number of samples. One way of achieving meaningful improvements on generic datasets is to pre-process the low resolution image with a functional model and add the (coarse) semantic results to the GAN, using a Conditional GAN (cGAN) framework.

We feel that these results reveal the current potential of these methods. Overall, the results are encouraging, as the network is capable of recovering details to improve the interpretability of the image, not just for improved quality purposes but for functionality of the imagery and potential impact on automatic target recognition (ATR) applications. As we enter an era where data quality becomes as important, if not more important, as the algorithm design, there is a challenge ahead of optimizing the methodology of how to train these networks for super-resolution tasks, as well as understanding limitations, expectations, and feasibility to recover information loss.

\section{Conclusion}
Given recent progress in super-resolution using Deep Learning, overhead imagery is one field that can leverage these advances and use it for improved information exploitation capabilities. In this exploratory work, we have explored state-of-the-art super-resolution neural network models and proposed an architecture based on dense blocks to carry out the task of increasing semantic meaning by adding plausible realistic information in the scene. Our model aims at learning the probability distribution mapping from the low-resolution to high-resolution using adversarial training from many exemplar data sets and then applying it to never before seen data. We have compared several generative models and architectures with several publicly available satellite and airborne image datasets (panchromatic electro-optical (EO) and multi-band images), and have shown what is realistically possible with today’s tools. We have also shown that a GAN framework with a collection of modified dense blocks in the generator can outperform state-of-the-art models that have been proposed for natural images.

 



{\small
\bibliographystyle{ieee}
\bibliography{egbib}

\begin{thebibliography}{10}\itemsep=-1pt

\bibitem{iarpamvs}
M.~Bosch, Z.~Kurtz, S.~Hagstrom, and M.~Brown.
\newblock A multiple view stereo benchmark for satellite imagery.
\newblock In {\em Proceedings of the Applied Imagery Pattern Recognition
  Workshop (AIPR)}, Washington, DC, USA, 2016.

\bibitem{bruna}
J.~Bruna, P.~Sprechmann, and Y.~LeCun.
\newblock Super-resolution with deep convolutional sufficient statistics.
\newblock In {\em Proceedings of the International on Learning Representations
  (ICLR)}, 2016.

\bibitem{spacenet}
CosmiQWorks, DigitalGlobe, and NVIDIA.
\newblock Spacenet.
\newblock http://explore.digitalglobe.com/spacenet, 2016.

\bibitem{pixelsrnn}
R.~Dahl, M.~Norouzi, and J.~Shlens.
\newblock Pixel recursive super resolution.
\newblock {\em CoRR}, abs/1702.00783, 2017.

\bibitem{lapgan-denton}
E.~Denton, S.~Chintala, A.~Szlam, and R.~Fergus.
\newblock Deep generative image models using a laplacian pyramid of adversarial
  networks.
\newblock In {\em Proceedings of the Conference on Neural Information
  Processing Systems (NIPS)}, Montreal, Canada, 2015.

\bibitem{srcnn}
C.~Dong, C.~C. Loy, K.~He, and X.~Tang.
\newblock Image super-resolution using deep convolutional networks.
\newblock {\em IEEE Transactions on Pattern Analysis and Machine Intelligence},
  38(2):295--307, 2016.

\bibitem{D-bib:freeman3}
W.~Freeman, T.~Jones, and E.~Pasztor.
\newblock Example based super-resolution.
\newblock {\em Proceedings of the IEEE Computer Graphics and Applications},
  22(2):56--65, 2002.

\bibitem{D-bib:freeman2}
W.~Freeman and E.~Pasztor.
\newblock Markov networks for superresolution.
\newblock In {\em Proceedings of the Annual Conference on Information Sciences
  and Systems (CISS)}, pages 1841--1848, 2000.

\bibitem{david-gpusrez}
D.~Garcia.
\newblock Image super-resolution through deep learning.
\newblock https://github.com/david-gpu/srez, 2016.

\bibitem{dcgan}
I.~Goodfellow, J.~Pouget-Abadie, M.~Mirza, B.~Xu, D.~Warde-Farley, S.~Ozair,
  A.~Courville, and Y.~Bengio.
\newblock Generative adversarial nets.
\newblock In {\em Advances in Neural Information Processing Systems (NIPS)},
  pages 2672--2680, 2014.

\bibitem{resnet}
K.~He, X.~Zhang, S.~Ren, , and J.~Sun.
\newblock Deep residual learning for image recognition.
\newblock In {\em Proceedings of the IEEE Conference on Computer Vision and
  Pattern Recognition (CVPR)}, Las Vegas, NV, USA, 2016.

\bibitem{densenet}
G.~Huang, Z.~Liu, and K.~Q. Weinberger.
\newblock Densely connected convolutional networks.
\newblock In {\em Proceedings of the IEEE Conference on Computer Vision and
  Pattern Recognition (CVPR)}, Honolulu, HI, USA, 2017.

\bibitem{pix2pix}
P.~Isola, J.~Zhu, T.~Zhou, and A.~A. Efros.
\newblock Deep residual learning for image recognition.
\newblock In {\em Proceedings of the IEEE Conference on Computer Vision and
  Pattern Recognition (CVPR)}, Honolulu, HI, USA, 2017.

\bibitem{keys}
R.~Keys.
\newblock Cubic convolution interpolation for digital image processing.
\newblock {\em IEEE Transactions on Acoustics, Speech, and Signal Processing},
  29(6):1153--1160, 1981.

\bibitem{drcn}
J.~Kim, J.~K. Lee, and K.~M. Lee.
\newblock Deeply-recursive convolutional network for image super-resolution.
\newblock In {\em Proceedings of the IEEE Conference on Computer Vision and
  Pattern Recognition (CVPR)}, Las Vegas, NV, USA, 2016.

\bibitem{laplacian}
W.~Lai, J.~Huang, N.~Ahuja, and M.~Yang.
\newblock Deep laplacian pyramid networks for fast and accurate
  super-resolution.
\newblock In {\em Proceedings of the IEEE Conference on Computer Vision and
  Pattern Recognition (CVPR)}, Honolulu, HI, USA, 2017.

\bibitem{srgan}
C.~Ledig, L.~Theis, F.~Huszar, J.~Caballero, A.~P. Aitken, A.~Tejani, J.~Totz,
  Z.~Wang, and W.~Shi.
\newblock Photo-realistic single image super-resolution using a generative
  adversarial network.
\newblock In {\em Proceedings of the IEEE Conference on Computer Vision and
  Pattern Recognition (CVPR)}, Honolulu, HI, USA, 2017.

\bibitem{vedai}
S.~Razakarivony and F.~Jurie.
\newblock Vehicle detection in aerial imagery.
\newblock {\em Journal on Visual Communication and Image Representation},
  34(C):187--203, Jan. 2016.

\bibitem{schutler}
S.~Schulter, C.~Leistner, and H.~Bischof.
\newblock Fast and accurate image upscaling with super-resolution forests.
\newblock In {\em Proceedings of the IEEE Conference Computer Vision and
  Pattern Recognition (CVPR)}, Boston,USA, 2015.

\bibitem{fcn}
E.~Shelhamer, J.~Long, and T.~Darrell.
\newblock Fully convolutional networks for semantic segmentation.
\newblock {\em CoRR}, abs/1605.06211, 2016.

\bibitem{tobias}
J.~T. Springenberg, A.~Dosovitskiy, T.~Brox, and M.~A. Riedmiller.
\newblock Striving for simplicity: The all convolutional net.
\newblock {\em CoRR}, abs/1412.6806, 2014.

\bibitem{tai}
Y.~Tai, J.~Yang, and X.~Liu.
\newblock Image super-resolution via deep recursive residual network.
\newblock In {\em Proceedings of the IEEE Conference on Computer Vision and
  Pattern Recognition (CVPR)}, Honolulu, HI, USA, 2017.

\bibitem{yang}
J.~Yang, J.~Wright, T.~S. Huang, and Y.~Ma.
\newblock Image super-resolution via sparse representation.
\newblock {\em IEEE Transactions on Image Processing}, 19(11):2861--2873, 2010.

\bibitem{Zeyde}
R.~Zeyde, M.~Elad, and M.~Protter.
\newblock On single image scale-up using sparse-representations.
\newblock In {\em Proceedings of the International Conference on Curves and
  Surfaces}, pages 711--730, 2012.

\end{thebibliography}
}

\end{document}